\documentclass[twocolumn]{article}
\usepackage{authblk}
\usepackage{amsmath}
\usepackage{amssymb}
\usepackage{mathtools}
\usepackage{blindtext} 
\usepackage[numbers,sort]{natbib}
\usepackage{url}
\usepackage[super]{nth}
\usepackage{filecontents}

\begin{document}

\date{2017 March 6}

\title{Application of a Shallow Neural Network to Short-Term Stock Trading}

\author[1]{Abhinav Madahar}
\author[2]{Yuze Ma}
\author[3]{Kunal Patel}
\affil[1]{Rutgers University, \textit {abhinav.madahar@rutgers.edu}}
\affil[2]{Rensselaer Polytechnic Institute, \textit {may7@rpi.edu}}
\affil[3]{Cherokee High School, \textit {kunalgpatel00@gmail.com}}

\maketitle

\begin{abstract}
Machine learning is increasingly prevalent in stock market trading. Though neural networks have seen success in computer vision and natural language processing, they have not been as useful in stock market trading. To demonstrate the applicability of a neural network in stock trading, we made a single-layer neural network that recommends buying or selling shares of a stock by comparing the highest high of 10 consecutive days with that of the next 10 days, a process repeated for the stock's year-long historical data. A $\chi^2$ analysis found that the neural network can accurately and appropriately decide whether to buy or sell shares for a given stock, showing that a neural network can make simple decisions about the stock market.
\end{abstract}

\section{Intro}
\bibliographystyle{unsrtnat}

Large companies (e.g. Google, Exxon Mobil, etc.) are not owned by a single person or a private group of individuals. Rather, these companies are split up into small pieces (shares) which are then sold to any individual who can afford them. The value of a single share depends on many factors, primarily its current and expected future profits. For example, a share in Apple (the largest company by total share value) is sold at approximately \$140 \cite{yahooapple}. As a company grows, its shares increase in value; Amazon shares sold for \$18 each in 1997 \cite{investopedia}, but are now worth over \$800 because of Amazon's massive growth since \cite{yahooamazon}. A company's shares are refered to as its "stock". Stock investors try to buy shares in companies that will grow dramatically; an investor who bought shares of Amazon in 1997 would now be very rich.

Many artificial intelligence (AI) techniques are applied to stock trading, such as genetic algorithms and probabilistic logic network \cite{wired}. Neural networks, an AI technique that simulates the human brain, have been applied to stock trading, but have seen limited success because they work best for systems with clear patterns, which stock markets do not have \cite{hurwitz}.

In a neural network, neurons are grouped into layers, which are combined to form a network that simulates a brain. A single neuron takes in many inputs from the neurons of the previous layer, multiples each by a weight, sums the products, adds a constant bias, and finally runs that sum through an activation function, often the sigmoid or hyperbolic tangent function. If we represent the inputs as a vector $x$, the weights as a vector $w$, the bias as a scalar $b$, the activation function as $f$, and the final output as a scalar $a$, then

$$a = f ( x \cdot w + b )$$

A neuron accepts a vector of the previous layer's values and outputs a scalar value, which is then sent to the next layer. By tuning the weights and biases, the neural network can learn how to interpret given data, a process called "training." A deep neural network has many layers, allowing it to find deeper patterns, leading to a dramatic rise in usage in the past few years, especially for tasks like image recognition and natural language processing \cite{wired}. A shallow neural network is a neural network with only 1 or a handful of layers, making it simpler but less powerful.

This study explores the utility of a shallow neural network on stock trading, specifically on deciding whether to buy or sell shares of a given company when given only stock information on said company.

\section{Methods \& Procedure}
\subsection{Experimental Trial}
The neural network was created in Python 2.7 using Numpy, but a similar design in another language or with other tools is expected to yield similar results.

From Yahoo Finance, the daily highs over the past year from for stocks in the Standards \& Poors 500 (S\&P 500) were downloaded. These highs were then grouped into $n$ chunks, each 10 days long.

The neural network takes a chunk as a 10-vector input. It has only 1 hidden layer, which has 20 neurons. It outputs a 2-vector of the form $(a, b)$. $a$ and $b$ are both recommendations in the interval $[0, 1]$ with 1 being a strong recommendation. $a$ recommends selling shares and $b$ recommends buying shares. The recommendation with the larger magnitude is used. For simplicity, the network's output $(a, b)$ is used to generate a Boolean $c$ equal to the index of the larger component. If the two recommendations are equally strong, then buying more shares is suggested.

$$a > b \implies c = 0$$
$$a \ngtr b \implies c = 1$$
$$\therefore$$
$$c = a \ngtr b$$

It is trivially shown that $c$ represents the recommendation to buy shares. If the network recommends buying, $c=1$. If the network recommends selling (i.e. not buying), then $c=0$.

To generate the training data, all but the most recent chunk were considered. To generate the $k^{th}$ training datum $(x_k, y_k)$, let $x_k$ be the $k^{th}$ 10-vector chunk and $y_k$ be the correct value of $c$ for said chunk. To calculate $y_k$, compare the highest high $H_k$ of the $k^{th}$ chunk with $H_{k+1}$, that of the $(k+1)^{th}$ chunk. If the price rises (i.e. $H_{k+1} > H_k$), then shares shought be bought (i.e. $y_k=1$). If the price does not rise, then shares should be sold (i.e. $y_k=0$). To summarize,
$$H_{k+1} > H_k \implies y_k = 1$$
$$H' \ngtr H_k \implies y_k = 0$$
$$\therefore$$
$$y_k = H_{k+1} > H_k$$

It is trivially shown that
$$y_k = max(x_{k+1}) > max(x_k)$$

The training data
$$\{(x_1, y_1), (x_2, y_2), (x_3, y_3), ..., (x_{n-1}, y_{n-1})\}$$
ignores the most recent chunk $(x_n, y_n)$ so that it can be used to test the neural network.
red
The neural network then trains on the training data using standard backpropagation and gradient descent. To test it, the network makes its decision $c_{n-1}$ on the second-most recent chunk $x_{n-1}$. The decision's accuracy (i.e. $c_{n-1} = y_{n-1}$) was then recorded.

Of the 500 stocks in the S\&P 500 available, 385 stocks were given to the neural network, which trained and tested itself over each.

\subsection{Control Trial}
To know if the neural network is useful, a control that randomly recommends buying or selling was applied to the S\&P 500 data. It generated a $c_{n-1}$ that was equally likely to be $0$ or $1$. $y_{n-1} = max(x_n) > max(n_{n-1})$ as in the experimental group. This decision's accuracy (i.e. $c_{n-1} = y_{n-1}$) was then recorded.

\section{Results}
The null hypothesis argues that the neural network knows nothing of the stocks on which it makes decisions and thus is no better than a random suggestion of either buying or selling.

Of the 500 stocks given to the control, 278 were decided correctly and 222 were decided incorrectly. Assume that the control's results are not statistically significant. If so, then a chi-squared analysis should show that the control is not statistically significant. Applying the standard formula,

$$\chi^2 \equiv \sum_{i} \frac{(O_i - E_i)^2}{E_i}$$
$$\chi^2 = \frac{(278 - 250)^2}{250} + \frac{(222 - 250)^2}{250}$$
$$\chi^2 = 6.272$$

Because there are two possible classifications of a result, correct or incorrect, there are 2 degrees of freedom. Using a significance level $\alpha = 0.01$ with a single degree of freedom would require $\chi^2 > 6.63$, which is not the case for the control group.

In the experimental trial, the neural network was given 385 stocks. If the null hypothesis is true, then 192.5 decisions would be correct and 192.5 would be incorrect. The neural network made an incorrect decision for 57 stocks and a correct decision for the remaining 328. Applying chi-squared,

$$\chi^2 \equiv \sum_{i} \frac{(O_i - E_i)^2}{E_i}$$
$$\chi^2 = \frac{(328 - 192.5)^2}{192.5} + \frac{(57 - 192.5)^2}{192.5}$$
$$\chi^2 = 190.755844$$

190.755844 is significantly higher than the required 6.63, so the null hypothesis is rejected, showing that the neural network has learned from the stock data and can make an informed decision based on it.

\section{Conclusions \& Discussion}
Neural networks have seen little success in stock-trading due to the lack of an obvious pattern in the stock market. This particular experiment was designed with simplicity in mind. The neural network has been shown to accurately predict whether the highest value of a stock for 10 consecutive days is higher than the highest value of said stock for the next 10 days. However, this is not very applicable to real-world stock trading because the network does not determine what the next high is, so the stock trader does not know when \textit{exactly} to buy or sell shares in the stock, only that he or she should at some point.

Changes to the configurations of the neural network, such as its dimensions and learning rate, may improve the neural network's accuracy. Improvements may also be found by an alternative design, such as a convolution neural network, or a significant alteration to the existing network, such as a deep neural network.

\section{Acknowledgements}
The authors thank hackNYU for providing food, drink, lodging, internet access, and countless other utilities during the development of the neural network. We also thank Yahoo for freely distributing stock data. We also thank Rufus Pollock for distributing a list of the tickers of the S\&P 500 companies.

\bibliography{\jobname}

\end{document}